# Shifting from Discrete to Continuous Reference Data: QSM-Derived Horizontal Tree Biomass Distribution for Deep Learning Biomass Estimation

Nils Griese[1]*, Christoph Kleinn[1] and Nils Nölke[1]

[1]*Department of Forest Inventory and Remote Sensing, University of Göttingen, Göttingen, Germany*

*Corresponding author: Tel: +49 551 39-233472; Email: nils.griese@uni-goettingen.de

**Abstract**

Conventional modeling approaches for LiDAR-based above-ground biomass (AGB) estimation rely on discrete plot-level inventory aggregates. This methodology introduces boundary-effect uncertainties that may severely degrade model performance within small field plots. To solve this limitation, we evaluate a Horizontal Biomass Distribution (HBD) reference mapped continuously from Quantitative Structure Models (QSMs). We trained a sparse 3D U-Net on simulated broadleaved forest structures using three AGB reference types: a standard forest inventory (FI) plot-level aggregate, an edge-effect-free QSM plot-level aggregate, and a continuous HBD mapping. Evaluating training plot sizes scaling from 100 to 2500 m$^2$, QSM-based models systematically outperformed FI approaches at small plot sizes. Specifically, for 100 m$^2$ plots, the HBD reference reduced the relative root mean square error (RRMSE) by $16.84 \pm 4.37$ % and increased $R^2$ by $0.22 \pm 0.05$ against the FI baseline. By replacing plot level aggregates with HBDs as AGB reference, this methodology corrects for edge-effects and shows that using an HBD-based reference enhances model performance for small plot sizes.

## 1. Introduction

Forest ecosystems play a critical role in the global carbon cycle, acting as substantial sinks for atmospheric carbon dioxide. Consequently, the accurate quantification and monitoring of above-ground biomass (AGB) are essential for sustainable forest management, climate change mitigation strategies, and carbon reporting[1,2]. Since living woody biomass of accounts for most of terrestrial AGB[3], capturing it is one of the core objectives of contemporary national forest inventories[4–7]. Traditional forest inventories (FIs) usually provide precise estimates for large areas, owing to their large sample sizes. However, they are labor-intensive and costly, which restricts their temporal resolution. Therefore, remote sensing technologies have become crucial for scaling woody AGB estimates. Among sensor types used for woody AGB modeling, active sensors like Radio Detection and Ranging (RADAR) and Light Detection and Ranging (LiDAR) have shown their capability for structural characterization of forests due to its ability to penetrate the canopy and capture three-dimensional vertical structure unmatched by passive sensor types[8–11]. Within the LiDAR domain, platform selection involves a trade-off between spatial coverage and structural detail. While Aerial Laser Scanning (ALS) enables landscape-scale assessments, Terrestrial Laser Scanning (TLS) and Mobile Laser Scanning (MLS) provide highly detailed single-tree measurements but are limited in spatial extent. Bridging this gap, Unmanned Aerial Laser Scanning (ULS) has emerged as a cost-effective approach for capturing three-dimensional forest structure at the individual-tree level across medium-sized areas.

The most widely adopted method for LiDAR-based biomass modeling is the plot-level approach. In this approach, statistical models are developed to relate plot-level AGB to plot-level metrics derived from the LiDAR point cloud[10,12,13]. The used LiDAR metrics typically include manually defined statistical summaries of height distributions and canopy density, which are selected for their strong correlation with standing biomass[13–16]. Despite the success of the area-based approach, the reliance on manually engineered features can limit model generalizability and fail to capture complex spatial patterns within the canopy. Recent advancements in computer vision and deep learning have facilitated a shift towards data-driven modeling approaches that learn features directly from raw data. Specifically, sparse 3D Convolutional Neural Networks have demonstrated superior performance compared to conventional feature-based machine learning approaches and classical regression models for LiDAR-based AGB modeling[17]. These deep learning architectures can effectively process the voxelized 3D structure of forest stands, offering a more robust alternative for AGB estimation. However, the accuracy of even the most advanced deep learning models is fundamentally constrained by the quality and nature of the reference data used for training. In approaches which use a FI-based reference, the reference AGB is a plot-level aggregate, which is either assumed to be uniformly distributed over the plot area or simply assigned to the center point of the plot. However, this ignores the actual spatial variability of biomass within the plot and, more critically, suffers from edge effects where the canopy of trees rooted inside the plot extends outside, or conversely, trees rooted outside have canopy parts extending in[18,19]. These geometric discrepancies may introduce substantial noise and uncertainty into the training data, which then may propagate to the final model performance. This effect is particularly strong for plot sizes which are small relative to the observed tree sizes. Additionally, forest structure plays an important role in this dynamic, where the effect of plot size is smaller for homogeneous forest structure e.g. in plantations, as compared to inhomogeneous natural forests[20,21]. To address these limitations prior studies suggest that larger plots sizes should be chosen for remote sensing-based biomass modeling, even though this drives the cost of the field data acquisition up[22–26].

As an alternative, recent research has proposed shifting from discrete, plot-level aggregates to a continuous representation of biomass known as Horizontal Biomass Distribution (HBD)[18,27,28]. The HBD concept posits that biomass should be considered continuously distributed within a plot, consistent with the physical distribution of tree components (branches, leaves, stem) rather than assigning all biomass to the stem locations alone. A previous simulation study of design-based estimation of AGB indicates that adopting an HBD-based reference holds significant potential for enhancing the precision of biomass mapping by mitigating the boundary effects inherent in conventional forest inventories[18]. To be able to access the HBD of trees, Quantitative Structure Models (QSM) can be used to derive a robust volumetric reference based on high resolution point clouds. A QSM is a hierarchically ordered geometric reconstruction of above-ground tree woody volumes as a set of 3D cylinders, which are fitted to an underlying point cloud[29–31]. In the past they have been proven to be a reliable non-destructive way to access single tree volume or AGB[29,32,33] and can be used to derive allometric models[34,35], enhance tree species classification[36], or derive species-specific branch biomass and then HBDs[27].

In this study, we investigate the impact of reference types (FI-based, QSM-based, HBD-based) on the performance of deep-regression AGB models. We utilize semi-synthetic data, combining real ULS single-tree point clouds and their corresponding TLS-based QSMs with synthetic tree positions and real forest inventory data, to compare a conventional FI-based reference against an HBD-based reference. To this end, we employ the BioDiv-3DTrees dataset[37] to simulate synthetic broadleaved forest stands. We train a sparse 3D-UNet on three distinct reference types:

(1) the conventional FI-based reference, fitting the model to FI-style plot-level aggregates, (2) the corresponding edge effect-free QSM-based plot-level reference, and (3) the continuous HBD derived from QSMs of the corresponding trees, fitting the model at subtree level.

We evaluate the impact of these reference types across different plot sizes used for training (100 m², 400 m², 900 m², 1600 m², 2500 m²) and for the HBD-based models, spatial resolutions of the final product (0.2 m, 0.5 m, 1 m, 2 m, 5 m) to derive implications for future LiDAR-based biomass modeling.

## 2. Methods

### 2.1 Data sources

The data basis for this study is the Biodiv-3DTrees dataset[37]. The dataset includes 3,386 species-labeled Quantitative Structure Models (QSMs) of broadleaved trees, which were used as a volumetric biomass reference in this study, and their corresponding ULS point clouds. To ensure reference data quality, the QSMs were filtered using the validation labels provided with Biodiv-3DTrees: only trees whose QSM-derived attributes fell within ±2 cm of TLS-measured diameter at breast height (DBH), ±1 m of TLS-measured tree height, and ±2 m² of TLS-measured crown projection area were retained. After filtering, 1889 trees remained for use in this study.

### 2.2 QSM-based HBD maps

Woody aboveground biomass (AGB) was derived from the QSMs by multiplying individual cylinder volumes with species-specific and diameter-dependent wood densities[38,39]. The resulting cylinder-level biomass values were vertically integrated using Monte-Carlo integration with $10^7$ points per tree at a spatial resolution of 10 cm, resulting in a Horizontal Biomass Distribution (HBD) map for each tree.

### 2.3 Forest scene generation

The forest scene generation was designed to produce a controlled and consistent training dataset which allows evaluating the impact of reference types, spatial resolution, and plot sizes on model performance. The training data should represent a fixed parameter across all experiments.

Prior to forest scene generation, individual trees were partitioned into training, validation, and test sets using a nested 85/15 split, yielding a final distribution of 72% training, 13% validation, and 15% test trees. Therefore, trees used for training were never used to derive the final performance metrics.

Square plots of five different sizes,100 m², 400 m², 900 m², 1600 m², and 2500 m², were generated for each split. Stem locations within each plot were drawn from a Poisson point process with a minimum stem-to-stem distance of 3 m, to prevent overlapping stems. The DBH range was bounded by the Biodiv-3DTrees dataset (7–87 cm). The DBH distribution for each scene was produced by combining randomly sampled stand-level parameters. Stem density was sampled uniformly between 250 and 600 stems per hectare. A randomly sampled target Reineke Stand Density Index (SDI) of 480 to 600 was then used to estimate the quadratic mean diameter (using a 25 cm reference diameter)[40]. The SDI is a metric of forest stocking that expresses the relative density of a stand based on the relationship between stand density and their quadratic mean diameter Dq, ensuring the simulated stands remain within realistic stocking limits. Finally, the estimated Dq was used as a scale parameter for a truncated lognormal DBH distribution within the dataset's DBH range. Combining a wide stem density range with a varying SDI yields a diverse dataset spanning from young forests to old-growth-style stands.

Plots were initially simulated with an additional 20 m outer buffer and cropped afterward. This ensured realistic edge effects by allowing crowns of trees where the stem is located outside the forest inventory plot to extend into the plot area and vice versa. For each tree in a generated scene, one of the 10 closest trees in DBH was randomly selected from the dataset and placed at the designated position. As a means of data augmentation, the corresponding ULS point cloud and HBD map were randomly rotated and flipped to introduce variability in the training data.

For the test datasets, real stem positions and DBH values from the Biodiversity Exploratories EP and FOX plot forest inventories were used, ensuring that model evaluation reflects realistic stand configurations despite synthetic training data.

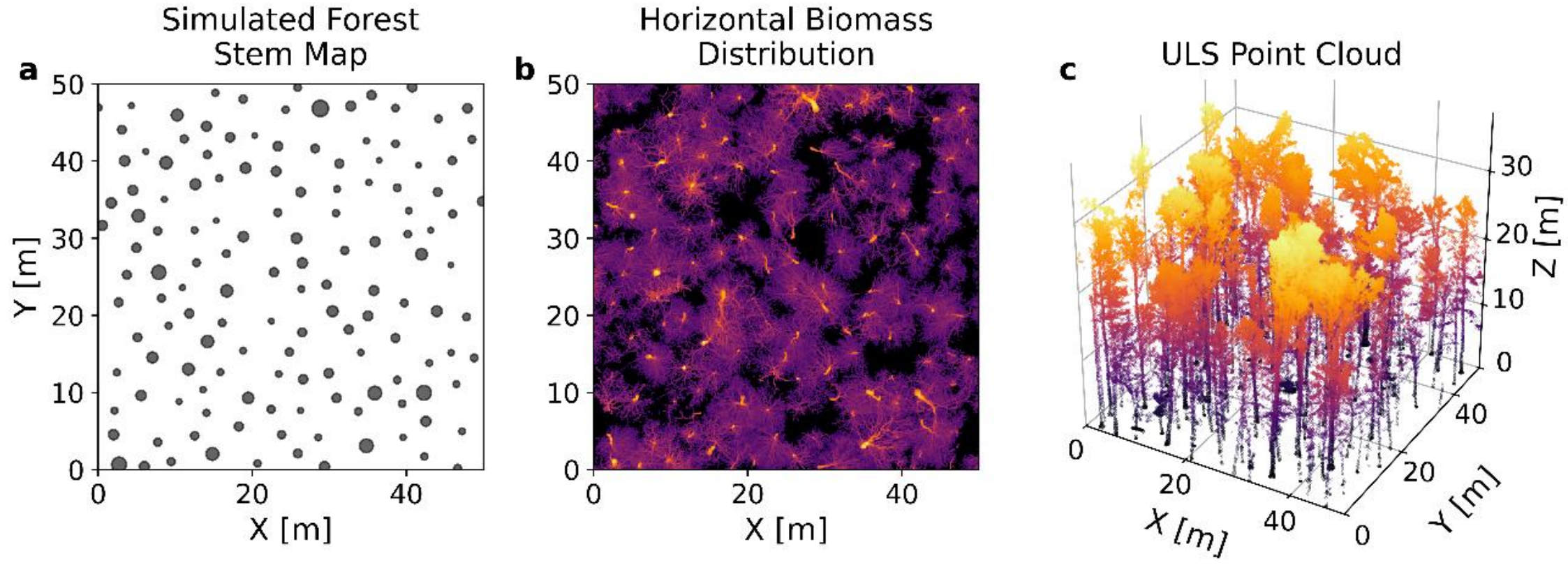

**Figure 1:** Example output of a generated forest plot. An initially generated forest inventory-style stem map (**a**) was used to distribute the single tree biomass to create a horizontal biomass distribution map (**b**). Corresponding single tree ULS point clouds were combined to create the forests ULS point cloud (**c**). Point size in (**a**) is according to the simulated DBH.

The forest generation produces three outputs per scene (see Figure 1): (1) a forest inventory stem map containing x and y coordinates, DBH, and QSM-based tree biomass for each stem; (2) the QSM-based HBD map of the plot; and (3) the corresponding ULS point cloud.

The total training data area per plot size was fixed to 450,000 m², which inversely scales the number of training plots with plot size (e.g., 4,500 plots of 100 m² vs. 180 plots of 2,500 m²). This reflects a realistic field campaign constraint, where a limited budget can be allocated to either a larger number of small plots or a smaller number of large plots. For the validation and test dataset 250 plots were generated for each plot size.

2.4 Model architecture

HBDNet is a sparse 3D U-Net that predicts HBD maps from ULS point clouds. The architecture is based on a modified version of TreeLearn[41], a single tree point cloud segmentation pipeline. For simplicity, we will focus on the modifications we made to the original TreeLearn architecture. For specific details about the model architecture please refer to the original paper[41].

For HBDNet, we replaced the tree-segmentation head with our regression head and adjusted the depth and resolution of the model. As TreeLearn, HBDNet is based on the concept of sparse convolutions[42], reducing the computational demand significantly. The model converts a ULS point cloud into a sparse voxel representation, processes it with sparse 3D convolutions, and projects the learned 3D feature-space into a two-dimensional biomass density map (see Figure 2).

In the first step, the input point cloud is voxelized to a resolution of 20 cm. For each occupied voxel, two attributes are computed: the maximum-normalized mean return intensity and the number of points falling within the voxel. We increased the voxel size to 20 cm from 10 cm to reduce the computational demand on our hardware. Additionally, we reduced the number of encoder and decoder layers of the U-Net to 5 from 7. We replaced the final normalization layer of TreeLearn with sparse submanifold convolution with kernel size 1 to get a single channel output and applied a softplus activation function to enforce non-negative model estimations. The sparse 3D output is densified and projected onto a two-dimensional horizontal grid by summing voxel values along the vertical axis, producing an HBD map at the native 0.2 m resolution. When a coarser output resolution is desired, the map is binned by summing within non-overlapping windows, preserving the total predicted biomass.

Two loss configurations were evaluated to compare the impact of a spatially explicit high-resolution biomass reference on model performance. The first configuration was used for the HBD-based models and combines a pixel-level Mean Absolute Error (L1 loss) with a plot-level L1 loss, encouraging global consistency alongside the pixelwise objective:

$$\mathcal{L}(y, \hat{y}) = L1(y, \hat{y}) + \frac{L1(\sum y, \sum \hat{y})}{n_{px}}$$

With $y$ and $\hat{y}$ being the ground truth and the predicted values, respectively; $\sum y$ and $\sum \hat{y}$ representing the plot level sums of prediction and ground truth, and $n_{px}$ representing the number of pixels at a given plot size and spatial resolution. The second configuration used for the Forest Inventory (FI)-based and plot-level QSM-based models based on the L1 Loss on the summed plot-level biomass values from the simulated forest inventory data and serves as a relative point of comparison, since this is the method generally used in previous LiDAR-based deep-regression AGB models[17,43].

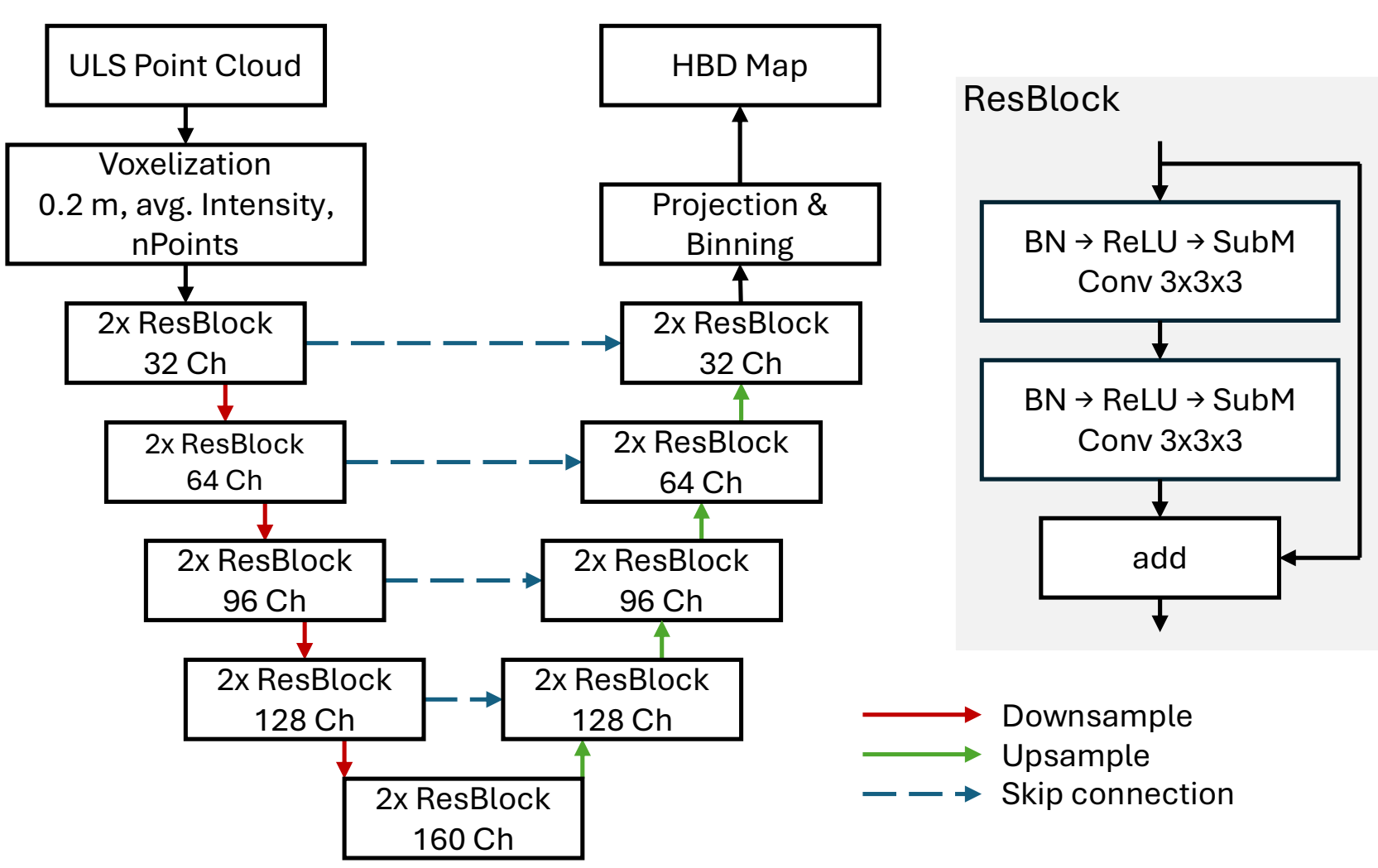

**Figure 2:** Schematic showing the overall architecture HBDNet (left) and the used ResBlock (right).

## 2.5 Model training

All code used in this study was written in Python. We used PyTorch[44], as well as the spconv library for the implementation of sparse convolutions[45]. Model training was done on Nvidia RTX 4080 and RTX 4090 GPUs. A five-fold cross-validation scheme was employed to average model performance on the test datasets and quantify prediction variability. We defined a constant random seed at the beginning of each training run, to minimize the impact of random model weight initializations. Models were trained separately for each combination of plot sizes and reference types. For models using HBD-based reference, we also changed the spatial resolution at which the model was optimized (20 cm, 50cm, 1 m, 2 m, 5 m).

## 2.6 Bias correction

Unlike traditional linear models, machine learning models do not automatically ensure that the sum of residuals is zero after training, meaning they are not inherently mean-unbiased. Therefore, we applied the method proposed by Igel and Oehmcke[46] to correct for this bias, which would otherwise accumulate at larger aggregation scales. This is achieved by adding a fixed correction offset to each model prediction, which is derived as the average residual on the training dataset after training.

## 2.7 Evaluation

Model performance was evaluated using real tree position maps from the Biodiversity Exploratories

Experimental Plots (EPs) and Forest Gap Experiments (FOX plots) and quantified using plot-level aggregates and the following metrics:

$$RMSE = \sqrt{\sum_{i=1}^{n} \frac{(\hat{y}_i - y_i)^2}{n}}$$

$$\text{RRMSE} = \sqrt{\frac{\frac{1}{n}\sum_{i=1}^{n}(y_i - \hat{y}_i)^2}{\sum_{i=1}^{n}(\hat{y}_i)^2}}$$

$$MAPE = \frac{100}{n}\sum_{i=1}^{n}\left|\frac{\hat{y}_i - y_i}{y_i}\right|$$

$$R^2 = 1 - \frac{\sum_{i=1}^{n}(y_i - \hat{y}_i)^2}{\sum_{i=1}^{n}(y_i - \frac{\sum_{i=1}^{n} y_i}{n})},$$

where $\hat{y}_i$is the reference value, $y_i$the model prediction, and $n$ the sample size. The variability between the cross-folds was quantified using the standard deviation of the respective performance metric:

$$\sigma = \sqrt{\frac{\sum_{i=1}^{k}(y_i - \bar{y})^2}{k-1}},$$

where $y_i$is the performance metric of a cross-fold, $\bar{y}$ is the average of the respective performance metric, and $k$ is the number of cross-folds (5 in this study).

**3. Results**

Figure 3 presents an example of a model prediction at a 20 cm spatial resolution on a plot from the test dataset. The predicted HBD map reproduced the main spatial biomass patterns of the reference distribution, including the location and extent of individual tree crowns. High-biomass regions were captured well spatially, although the prediction appears smoother than the reference Figure 4 illustrates the average model performance on the test datasets as a function of training plot size.

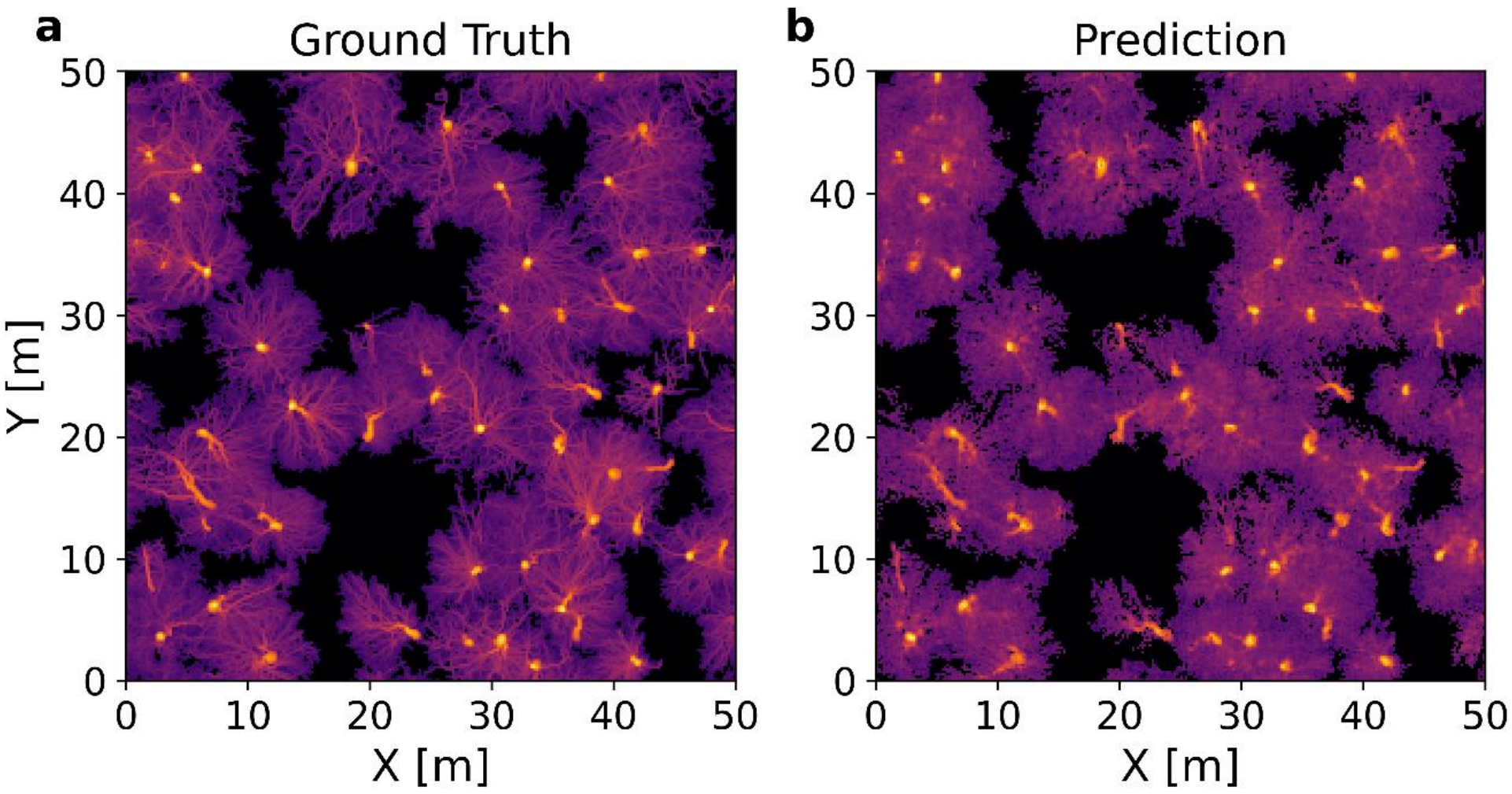

**Figure 3:** Example of an HBD-based model prediction (**b**) next to the HBD-based ground truth (**a**) for a 2500 m² plot at a spatial resolution of 20 cm. For visualization, the 8th root is shown in both cases.

Overall, models utilizing QSMs as reference data outperformed models fitted on discrete forest inventory data. The performance differences between forest inventory-based and QSM-based models were most pronounced at the smallest evaluated plot size of 100 m². Specifically, for models operating at a 1 m spatial resolution, the RMSE was reduced by up to 34.1 ± 9.84 Mg ha$^{-1}$, the RRMSE improved by 16.84% ± 4.37%, the $R^2$ increased by 0.22 ± 0.05, and the MAPE improved by 36.36% ± 20.64%. These performance deviations decreased as training plot sizes increased and became negligible at plot sizes larger than 900 m².

Furthermore, models fitted at using the HBD-based reference generally performed slightly better than those fitted at the plot level. This difference is especially noticeable at smaller plot sizes (≤ 400 m²; see Figure 5). For 100 m² plots and 1 m spatial resolution the RMSE was reduced by 12.32 Mg ha$^{-1}$ ± 7.45 Mg ha$^{-1}$, the RRMSE was reduced by 6.11% ± 3.65%, the $R^2$ increased by 0.092 ± 0.054, and the MAPE was reduced by 22.57% ± 15.24% compared to the plot-level QSM-based model. At larger plot sizes, the performance difference between HBD-based and plot-level fitted models is marginal at the plot-level scale. For the HBD-based models, higher spatial resolutions (≤ 2 m) achieved the best results. While the RMSE, RRMSE, and MAPE consistently decreased with increasing training plot sizes for both HBD-based and plot-level models, the $R^2$ metric showed a more nuanced picture. All model variants demonstrated a degree of $R^2$ saturation. However, models utilizing QSMs as reference data exhibited a shallower trajectory across training plot sizes. Conversely, models trained on forest inventory-style references did not plateau completely, even at the maximum evaluated plot size of 2500 m². Finally, HBD-based models attained higher $R^2$ values utilizing smaller training plots.

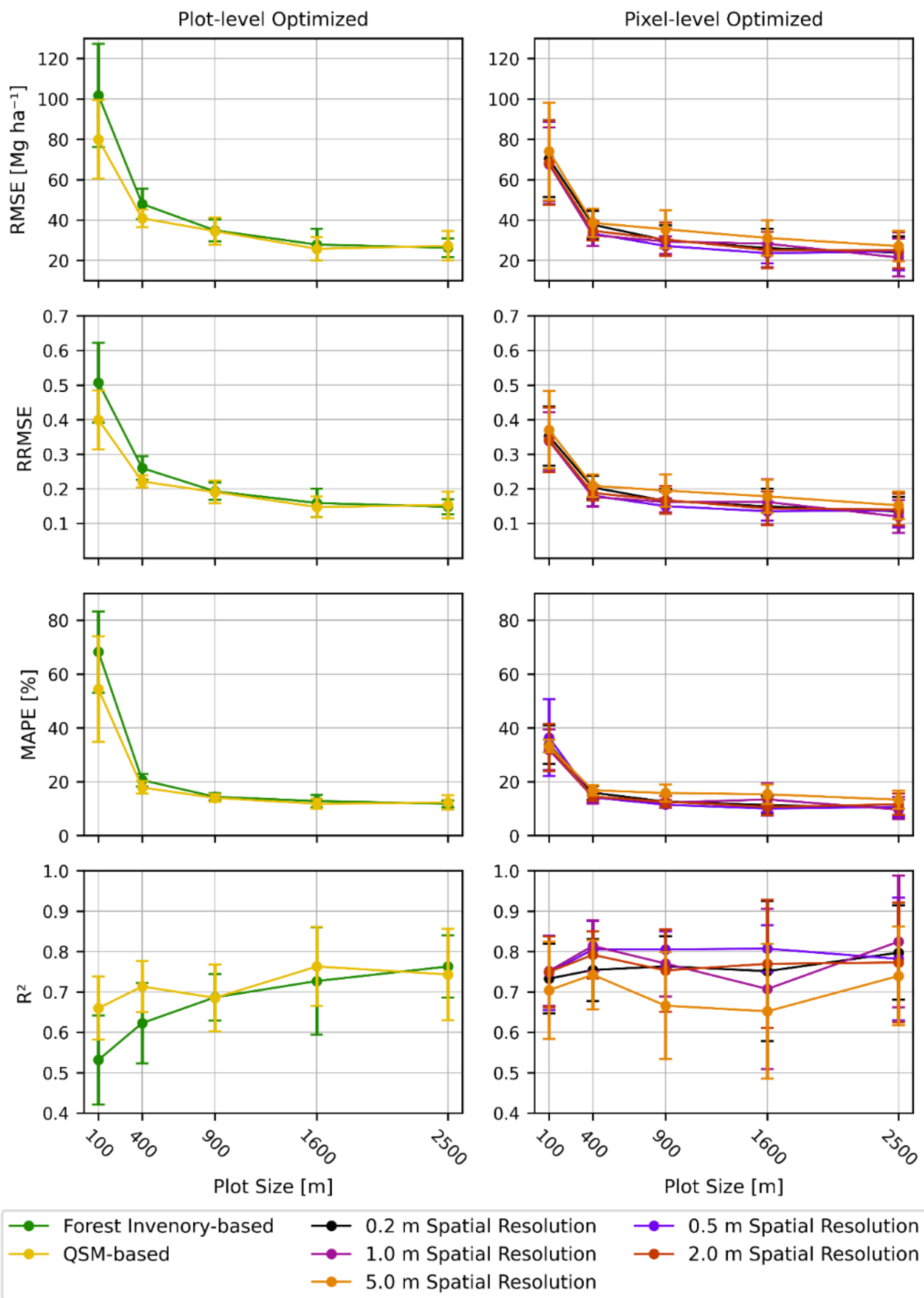

**Figure 4:** Average model performance on the test datasets of the cross folds (n=5) for plot-level fitted (left) and HBD-based models (right) at different training plot sizes and spatial resolutions. The error bars show one standard deviation of the metric values at each point.

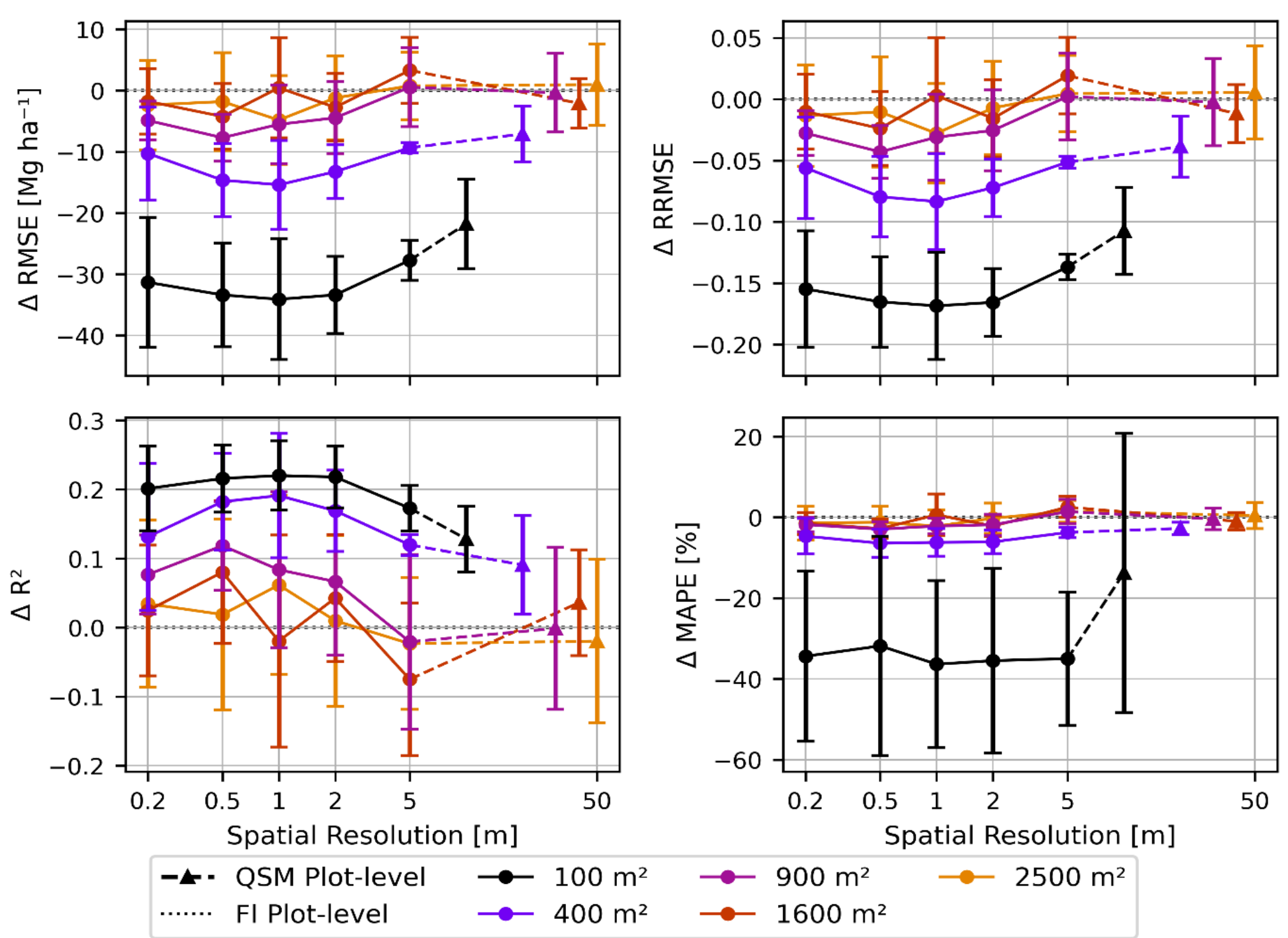

**Figure 5:** Comparison of model performance of each QSM-based model with the forest inventory-based model (dashed line at zero level) as the difference Δ in RMSE, RRMSE, $R^2$, and MAPE at different spatial resolutions used during training. The triangle markers furthest to the right show the performance using edge-effect free QSM-based plot-level aggregates.

## Discussion

### General Discussion of the Results

To our knowledge, this study presents the first approach to remote sensing-based modeling of woody above-ground biomass (AGB) utilizing the horizontal biomass distribution (HBD) of trees and forests. Unlike conventional approaches that rely on plot-level aggregates, we employed Quantitative Structure Models (QSMs) alongside species- and diameter-specific wood densities to derive high-resolution HBD maps, which served as our ground truth. Within the scope of the presented simulation framework, the increased complexity and detail of these ground truth measurements enhanced model performance, particularly when training on smaller plot sizes.

While the performance metrics presented in our study improved as training plot sizes increased, the distinct advantage of our methodology became particularly evident at smaller plot sizes. Here, the models using the plot-level QSM-based reference proved more capable at smaller plot sizes ($\leq 400\ \text{m}^2$) than FI-based approaches.

We assume that by providing the plot-level HBD as reference, boundary-related discrepancies were corrected, allowing a better model performance. This mechanism is supported by the fact that the models using the HBD-based reference also yielded notably better predictive performance than the FI-based reference. Both QSM-based references differ from the FI-based reference strictly in its geometrical treatment of the plot perimeter. They accurately exclude the biomass components of 'inside-trees' that extend beyond the boundaries, while simultaneously including the overhanging branches of 'outside-trees' that reach into the plot. The additional information about the intra-plot HBD further improved model performance, following the same pattern of larger performance gains at smaller plot sizes.

### Comparison with Other Studies

Direct comparisons with existing literature should be made with care, since in our study we used semi-synthetic forest stands assembled from real singletree point clouds. However, comparable simulation-based approaches exist[22,47]. Our overall trend that model performance increases with larger plot sizes aligns with the standard internal validation trends reported in the literature[22,48–50]. However, as demonstrated in a previous study[47], the apparent superiority of larger plots may be an artifact of internal validation frameworks. Models trained on smaller plots can perform as well as those trained on larger plots, provided they are not restricted by unequal sampling intensity or edge effects. By replacing discrete stem-based biomass assignments with QSM-derived HBD references, we correct for the edge effects that typically degrade model performance when using smaller reference areas (i.e., those $\leq 400\ \text{m}^2$).

Concerning deep regression approaches for AGB estimation, our findings build upon and contextualize recent methodological advancements. A previous study demonstrated the utility of sparse 3D U-Nets for biomass prediction[17], pairing real-world ALS point clouds with discrete National Forest Inventory aggregates. To address boundary inconsistencies, they incorporated custom edge-effect data augmentations, reporting a Mean Absolute Percentage Error (MAPE) of ≥ 84.42% on their test set. However, the expected differences between their real-world dataset and our simulation study do not allow for direct quantitative comparison. Nevertheless, we would argue that while augmentation strategies can reduce a model's sensitivity to edge artifacts, relying on FI aggregates can limit a model's performance.

Similarly, Pourdelan et al. emphasized the value of synthetic environments for training deep regression models, utilizing point-based architectures trained on simulated leaf-on Eucalyptus scans[43]. Because their reference data was aggregated from 3D mesh models, their target variable was inherently free of edge effects, conceptually mirroring the QSM-based plot-level reference evaluated in our study. This underscores the importance of using precise, geometrically consistent references for deep regression. Unlike traditional statistical models, deep learning models hold the capacity to fully exploit detailed references like QSMs or meshes, thereby leveraging information that is typically discarded in conventional plot-level aggregation.

Limitations

Since the presented approach relies on high-quality QSMs to link the reference AGB density and ULS point clouds at the intra-crown level, this study was restricted to broadleaved trees. QSM reconstruction quality for QSMs is frequently lower for coniferous species due to occlusion caused by needles[33,51].

A methodological limitation of the current model architecture originates from the use of sparse submanifold convolutions which cannot yield non-zero values in regions where no points from the input point cloud are present. This leads to a guaranteed underestimation of AGB within the interior of tree stems larger than the used voxel resolution (here 20 cm). We accounted for this limitation by adding a final binning operation, which sums up multiple pixels, to reduce the final resolution of the model output. While the models binned to 1 m spatial resolution yielded the highest performance in this study, future revisions of this approach should consider modifying the convolutional layers to allow for the generation of filled voxels, though this will likely increase the overall computational effort.

Currently, the proposed framework relies on high-quality QSM reconstructions for trees contributing biomass within and around the plot boundaries, which can require substantial data acquisition and post-processing effort. Future work should therefore investigate simplified or parameterized representations of horizontal biomass distribution that preserve the spatial characteristics of biomass allocation while reducing reference-data requirements[27]. Conversely, the operational burden of applying the presented framework may be reduced through the use of pre-trained model weights. Future users can leverage these weights to fine-tune the model, thereby reducing the volume of reference data required compared to training a network from scratch.

Conclusion

In this study we evaluated the potential of utilizing QSMs as continuous AGB reference for LiDAR-based AGB estimation. We found the largest gain from QSM-based AGB reference for small plot sizes (≤ 400 m²) and a spatial resolution of 1 m. Using the high-resolution HBD-based AGB reference further improved model performance at these small plot sizes. While further validation under fully real-world conditions is required, the study demonstrates the potential of continuous QSM-derived biomass references for future LiDAR-based AGB modeling workflows.

**Acknowledgements**

Funding for Nils Griese was provided by the German Research Foundation (DFG project number 496533645).

**Author contributions**

Nils Griese performed the formal analysis, visualization, software development, investigation, data curation, and wrote the initial draft. Christoph Kleinn and Nils Nölke provided supervision, resources, project administration, and funding acquisition, and reviewed and edited the manuscript. All authors discussed the results and approved the final manuscript.

**Data availability**

The datasets analyzed during the current study are available in the GRO.data repository https://doi.org/10.25625/8PB1IF. The code used for model training and evaluation can be found at https://gitlab.gwdg.de/griese1/hbdnet.

**Competing interests**

The author(s) declare no competing interests.